\documentclass[runningheads]{llncs}

\usepackage[T1]{fontenc}
\usepackage{graphicx}
\usepackage{amsmath}
\usepackage{amssymb}
\usepackage{booktabs}
\usepackage{array}
\usepackage{float}
\usepackage{placeins}
\usepackage{url}
\usepackage{microtype}
\usepackage{marvosym}

\let\svthefootnote\thefootnote
\newcommand\freefootnote[1]{%
  \let\thefootnote\relax%
  \footnotetext{#1}%
  \let\thefootnote\svthefootnote%
}
\newcolumntype{L}[1]{>{\raggedright\arraybackslash}p{#1}}
\newcommand{\XLBaseSmallFOne}{0.367}
\newcommand{\XLFinalSmallFOne}{0.354}

\newcommand{\XLFinalDSC}{0.712}

\newcommand{\XLFinalNSD}{0.714}
\newcommand{\XLBaseAllFOne}{0.741}
\newcommand{\XLFinalAllFOne}{0.779}
\newcommand{\XLBaseHD}{69.2}
\newcommand{\XLFinalHD}{56.6}

\newcommand{\XLDeltaSmallFOne}{-0.021}
\newcommand{\XLDeltaSmallFOneLow}{-0.049}
\newcommand{\XLDeltaSmallFOneHigh}{0.003}
\newcommand{\XLDeltaDSC}{+0.044}
\newcommand{\XLDeltaDSCLow}{0.030}
\newcommand{\XLDeltaDSCHigh}{0.059}
\newcommand{\XLDeltaNSD}{+0.038}
\newcommand{\XLDeltaNSDLow}{0.023}
\newcommand{\XLDeltaNSDHigh}{0.054}
\newcommand{\XLDeltaAllFOne}{+0.019}
\newcommand{\XLDeltaAllFOneLow}{0.007}
\newcommand{\XLDeltaAllFOneHigh}{0.030}
\newcommand{\XLDeltaHD}{13.36}
\newcommand{\XLDeltaHDLow}{6.82}
\newcommand{\XLDeltaHDHigh}{20.43}

\newcommand{\XLETAllFOne}{0.847}
\newcommand{\XLETDSC}{0.753}
\newcommand{\XLETNSD}{0.810}
\newcommand{\XLRCAllFOne}{0.604}

\newcommand{\XLTCAllFOne}{0.845}
\newcommand{\XLTCDSC}{0.775}
\newcommand{\XLTCNSD}{0.821}
\newcommand{\XLWTAllFOne}{0.820}
\newcommand{\XLWTDSC}{0.732}
\newcommand{\XLWTNSD}{0.750}

\begin{document}
\raggedbottom

\title{Region-Wise Logit Fusion and Utility-Gated Refinement for Brain Metastasis Segmentation}
\titlerunning{COMPASS-METS for Brain Metastasis Segmentation}

\author{Zhenye Lou\inst{1} \and 
Qing Xu\inst{2}\textsuperscript{$^{\dag}$} \and
Wenting Duan\inst{3} \and
Zhen Chen\inst{1}\textsuperscript{(\Letter)}}


\authorrunning{Z. Lou et al.}

\institute{
Department of Data Science and Artificial Intelligence,\\The Hong Kong Polytechnic University, Hong Kong SAR \and
School of Computer Science, University of Nottingham, UK \and
School of Engineering and Physical Science, University of Lincoln, UK 
\\
\email{scxqx1@nottingham.ac.uk, z.chen@polyu.edu.hk}
}

\maketitle

\begin{abstract}
BraTS-METS 2026 Task~1 evaluates segmentation of brain metastases and postoperative resection cavities on heterogeneous multiparametric MRI, requiring detection of small enhancing foci while preserving the geometry and hierarchy of enhancing tumor (ET), tumor core (TC), whole tumor (WT), and resection cavity (RC). We present COMPASS-METS (Component-Aware Multi-Trainer Probability Assembly with Structured Segmentation), which combines three five-fold nnU-Net configurations into aligned regional probability ensembles. A structured-probability anchor applies region-specific component decisions, TC boundary completion, RC filtering, and hierarchy-preserving conversion. Omitted ET candidates are recovered additively using learned component confidence and deterministic multi-trainer support for ET and its TC/WT parents; a final utility gate admits only disconnected candidates with sufficient lesion-existence and geometry-safety probabilities. On the reused 179-case challenge development cohort, COMPASS-METS (Team: MicroBT) achieved Small F1 \XLFinalSmallFOne, lesion-wise DSC \XLFinalDSC, NSD \XLFinalNSD, All F1 \XLFinalAllFOne, and HD95 \XLFinalHD\,mm. Relative to ResEncXL, it improved DSC, NSD, and All F1 and reduced HD95, while Small F1 changed from \XLBaseSmallFOne{} to \XLFinalSmallFOne. Stagewise analysis attributes the principal geometric gains to structured probability construction and shows that guarded ET refinement recovers most of the associated small-lesion detection loss. These 179-case results remain selection-aware development estimates. \freefootnote{$^{\dag}$ Team Leader.}

\smallskip\noindent The source code is publicly available at
\url{https://github.com/leonlou921/COMPASS-METS}

\end{abstract}

\section{Introduction}

Automatic delineation of brain metastases can support volumetric assessment, treatment planning, and longitudinal monitoring \cite{karargyris2023federated,luo2025msm}. The task is difficult because a single examination may contain many spatially separated lesions, including very small enhancing foci, and post-treatment scans may additionally contain resection cavities (RC). Previous BraTS-METS analyses found lower detection and segmentation performance for decreasing lesion volume, particularly below $100\,\mathrm{mm}^3$ \cite{moawad2024brain,xu2023dcsau,liu2025harnessing}. BraTS-METS 2026 Task~1 directly evaluates this failure mode through lesion-level metrics and defines targets below $27\,\mathrm{mm}^3$ as small lesions \cite{https://doi.org/10.5281/zenodo.19714728}.

Methods for this problem have progressed from template-based detection to deep three-dimensional segmentation \cite{ambrosini2010computer,cho2021brain}. Modern self-configuring systems adapt well to heterogeneous cohorts, but their voxel-wise probabilities expose a detection--geometry trade-off at inference. Conservative component decisions suppress distant false positives and boundary outliers, yet they can also remove small, weakly supported true lesions. Relaxing a global threshold reverses that trade-off but may introduce geometrically costly false positives. The output semantics add another constraint: ET is contained in TC, TC in WT, and RC remains independent of the nested tumor regions.

\begin{figure}[t]
    \centering
    \includegraphics[width=0.85\linewidth]{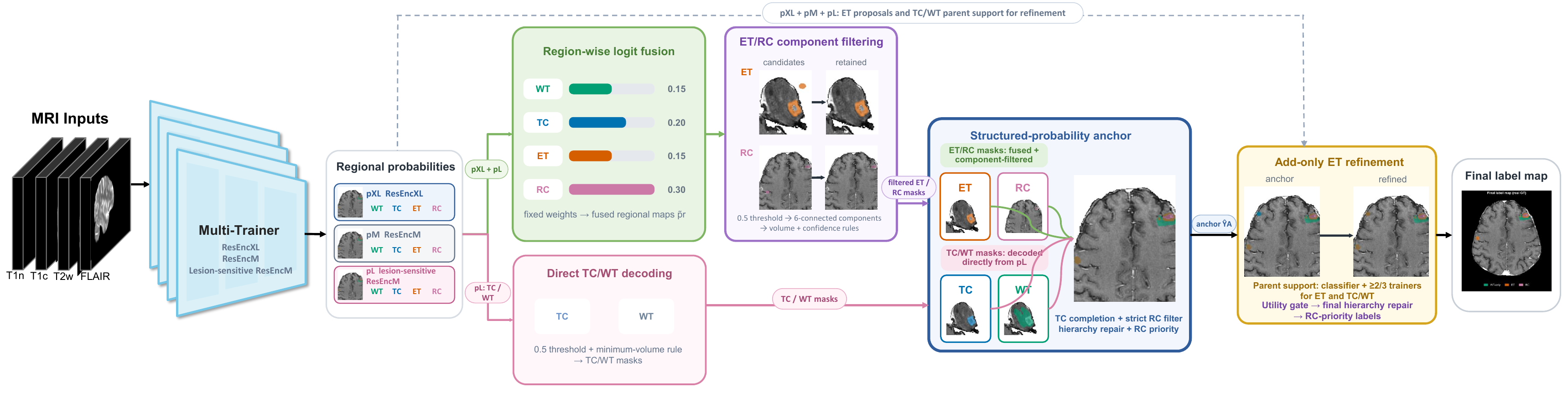}
    \caption{Overview of COMPASS-METS. Mirrored inference and five-fold averaging yield aligned regional probability sources $p^{\mathrm{XL}}$, $p^{\mathrm{M}}$, and $p^{\mathrm{L}}$. Region-wise logit fusion of $p^{\mathrm{XL}}$ and $p^{\mathrm{L}}$ provides ET/RC maps for component filtering, whereas $p^{\mathrm{L}}$ provides TC/WT masks directly; the two routes form the structured-probability anchor. All three probability sources then support add-only ET refinement, followed by utility gating, hierarchy repair, and RC-priority label conversion.}
    \label{fig:pipeline}
\end{figure}

We address this trade-off through multi-trainer probability construction and guarded component refinement (Fig.~\ref{fig:pipeline}). ResEncXL and ResEncM are optimized in five folds, and the lesion-sensitive ResEncM configuration is initialized from the fold-matched ResEncM checkpoint. Their aligned regional probabilities form a structured anchor and provide disagreement evidence for component-level decisions.

Our contributions are: (i) a multi-trainer route that preserves aligned soft probabilities until region-specific decisions; (ii) parent-supported ET refinement requiring agreement between a learned component model and ET/TC/WT probabilities; and (iii) a utility gate that models lesion existence and geometric safety separately. On the 179-case challenge development cohort, stagewise comparisons showed that structured probability construction produced the main geometric improvements, while parent-supported refinement recovered most of the associated loss in small-lesion detection.

\section{Data and Evaluation}

\begin{figure}[t]
    \centering
    \includegraphics[width=0.56\linewidth]{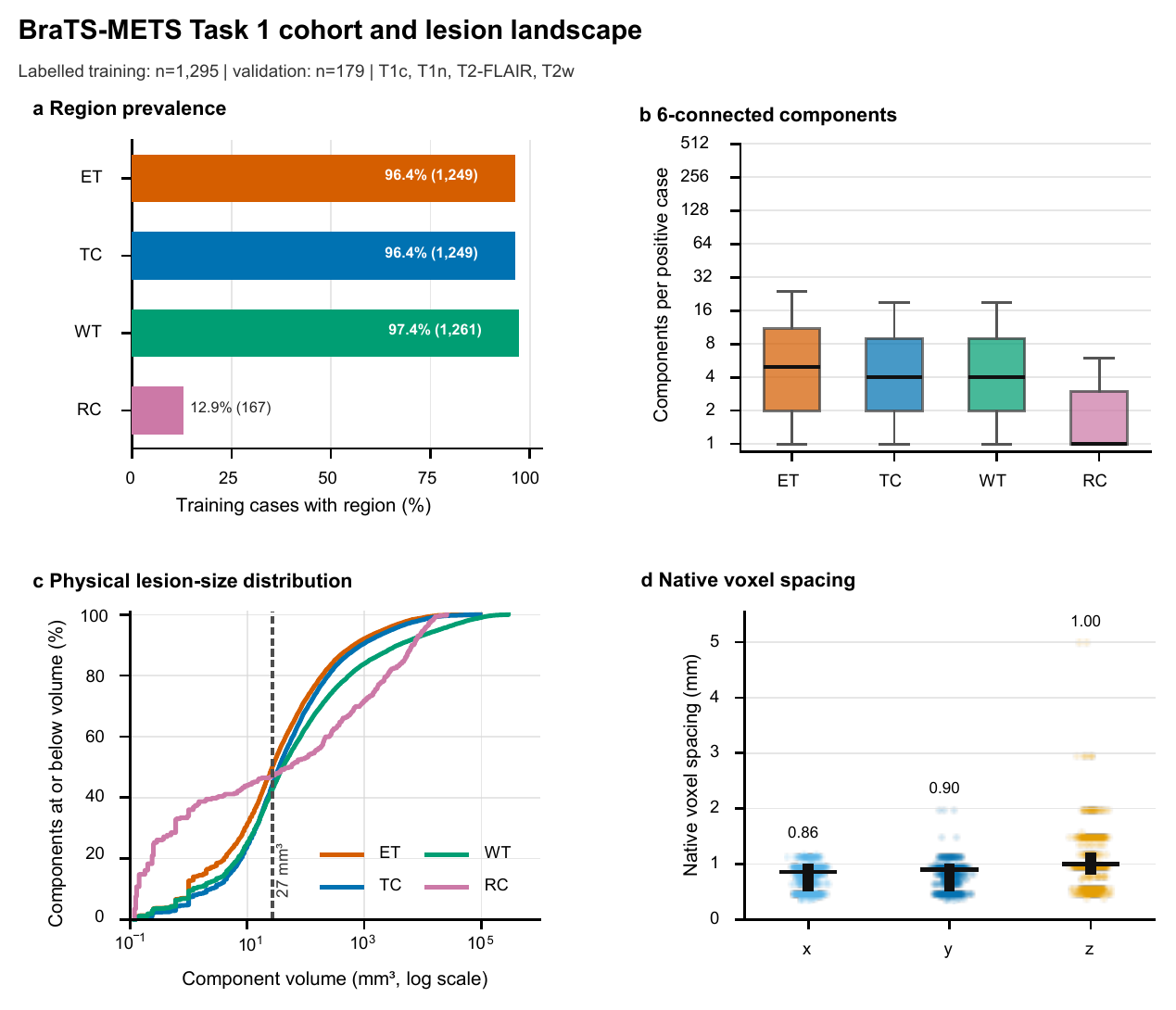}
    \caption{Analyzed training-cohort statistics: regional prevalence, 6-connected components per positive examination, physical component volumes, and native voxel spacing. The dashed guide marks $27\,\mathrm{mm}^3$.}
    \label{fig:dataset}
\end{figure}

The BraTS-METS release contains retrospective pre- and post-treatment multiparametric MRI acquired across institutions, scanners, and imaging protocols \cite{maleki2025analysis}. Each examination provides four co-registered channels: native T1-weighted (T1n), contrast-enhanced T1-weighted (T1c), T2-weighted FLAIR (T2-FLAIR), and T2-weighted (T2w) MRI; under the challenge protocol, the T2w channel may be native or synthetic \cite{https://doi.org/10.5281/zenodo.19714728}. Figure~\ref{fig:dataset} summarizes the resulting variation in regional prevalence, component volume, and native voxel spacing.

The analyzed release contains 1,295 annotated training examinations, each covered once across five held-out folds. The 179 hidden-label validation cases were scored through the challenge interface. Because this feedback informed trainer and inference selection, these results are selection-aware development estimates rather than an independent generalization test. Data were accessed through Synapse project syn74274097 under the challenge data-use terms \cite{https://doi.org/10.5281/zenodo.19714728}; no identifiable clinical information was exported by our pipeline.

The mutually exclusive labels are non-enhancing tumor core (NETC), surrounding non-enhancing FLAIR hyperintensity (SNFH), ET, and RC. Whole tumor (WT) is the union of NETC, SNFH, and ET; TC is the union of NETC and ET; ET and RC are evaluated separately. Tumor outputs are constrained by $\mathrm{ET}\subseteq\mathrm{TC}\subseteq\mathrm{WT}$, while RC remains outside this nesting.

Task~1 defines components below $27\,\mathrm{mm}^3$ as small targets and evaluates detection with lesion geometry \cite{https://doi.org/10.5281/zenodo.19714728}. Final estimates use the returned challenge score files. Candidate screening additionally decomposes each region into 6-connected components and greedily pairs overlapping components by descending DSC without an IoU threshold. Unmatched ground truth receives zero DSC/NSD and image-diagonal HD95; NSD tolerance is 1.0~mm. Empty references receive perfect geometry only for empty predictions.

Aggregate scores macro-average finite ET, RC, TC, and WT mean records. For the 179-case challenge analysis, paired intervals align subjects, macro-average finite regional case entries, and use 10,000 paired case-level bootstrap resamples with percentile 95\% intervals. Positive changes are favorable; HD95 is sign-inverted to express a reduction. Screening precision, recall, and F1 are micro-averaged from TP/FP/FN counts and use $275\,\mathrm{mm}^3$, whereas final analysis uses the official $27\,\mathrm{mm}^3$ threshold.

We additionally evaluate the complete pipeline and three pure ensembles on all 1,295 training cases using XL, M, and lesion-sensitive predictions from each assigned held-out fold. Equal-probability, equal-logit, and hard-majority controls use threshold 0.5 and only the common hierarchy decoder. Paired OOF intervals use 5,000 case-level bootstrap resamples. Learned models exclude the evaluated fold, but operating points remain global rather than inner-fold selected; this is a post-selection held-out-fold evaluation, not a fully nested estimate.

\section{Methods}

\subsection{Overview of the Proposed Method}

COMPASS-METS has four inference stages (Fig.~\ref{fig:pipeline}). First, the ResEncXL, ResEncM, and lesion-sensitive ResEncM trainers produce aligned five-fold probabilities for WT, TC, ET, and RC; the lesion-sensitive trainer is initialized from its fold-matched ResEncM checkpoint. Second, ResEncXL and lesion-sensitive probabilities form a structured-probability anchor. Third, a learned component model and two-of-three parent support recover ET proposals missed by the anchor. Finally, separate lesion-existence and geometry-safety models screen additional disconnected ET candidates. Every refinement is add-only, after which explicit set operations preserve $\mathrm{ET}\subseteq\mathrm{TC}\subseteq\mathrm{WT}$ and convert the regions to mutually exclusive labels.

\subsection{Multi-Trainer Training and Probability Construction}

The primary trainer uses the nnU-Net v2 residual-encoder XL full-resolution plan \cite{isensee2021nnu,isensee2023extending}. Automatically configured preprocessing determines spacing, resampling, cropping, and channel-wise normalization. Five folds were trained for 1000 epochs with $160\times224\times192$ patches, batch size 3, regional sigmoid outputs, and Dice--binary-cross-entropy loss. Stochastic-gradient descent used learning rate $10^{-2}$, momentum 0.99, Nesterov acceleration, weight decay $3\times10^{-5}$, and a polynomial schedule. Deep supervision and configured augmentations were retained; the checkpoint with the highest smoothed foreground pseudo-Dice was selected per fold. Two residual-encoder M trainers provide additional sources. The standard configuration uses the same regional objective. The lesion-sensitive configuration starts from the corresponding standard fold and adds focal and asymmetric Tversky terms, each weighted 0.5, with $(\alpha_F,\gamma)=(0.25,2)$ and $(\alpha_T,\beta_T)=(0.3,0.7)$. All five lesion-sensitive folds are fine-tuned at $10^{-3}$ with checkpoint-best selection. Mirrored inference averages five outputs per trainer after alignment to the ResEncXL grid, producing $p^{\mathrm{XL}}$, $p^{\mathrm{M}}$, and $p^{\mathrm{L}}$. Small-lesion oversampling and synthetic-lesion augmentation were screened but not retained.

\subsection{Structured-Probability Anchor}

The structured anchor provides conservative geometry before any lesion recovery. To avoid infinite logits, define $p^\epsilon=\min(\max(p,\epsilon),1-\epsilon)$ with $\epsilon=10^{-5}$. ResEncXL and lesion-sensitive probabilities are fused by region:
\begin{equation}
\widetilde p_r(x)=\sigma\!\left[(1-w_r)\operatorname{logit}p^\epsilon_{\mathrm{P},r}(x)
+w_r\operatorname{logit}p^\epsilon_{\mathrm{A},r}(x)\right].
\label{eq:logitfusion}
\end{equation}
Here $p_{\mathrm{P}}=p^{\mathrm{XL}}$, $p_{\mathrm{A}}=p^{\mathrm{L}}$, and $(w_{\mathrm{WT}},w_{\mathrm{TC}},w_{\mathrm{ET}},w_{\mathrm{RC}})=(0.15,0.20,0.15,0.30)$. This is logit fusion rather than calibration because no reliability objective is fitted. These weights and component operating points were inherited from ResEncM/lesion-sensitive development and transferred without a dense XL sweep; the final pipeline was nevertheless retained using 179-case feedback.

The anchor takes TC/WT from $p^{\mathrm{L}}$ and ET/RC from $\widetilde p$. Maps are thresholded at 0.5 and decomposed into 6-connected components. Given voxel volume $\nu$, minimum volume $v_r$, and mean/peak thresholds $(\mu_r,\pi_r)$, ET or RC component $C$ is retained when
\begin{equation}
G_r(C)=\mathbf 1\!\left[
|C|\nu\ge v_r,\quad
|C|^{-1}\!\sum_{x\in C}\widetilde p_r(x)\ge\mu_r,\quad
\max_{x\in C}\widetilde p_r(x)\ge\pi_r
\right].
\label{eq:component}
\end{equation}
ET uses $(v_r,\mu_r,\pi_r)=(5~\mathrm{mm}^3,0.55,0.70)$ and RC uses $(20~\mathrm{mm}^3,0.65,0.80)$. TC and WT use only a $10~\mathrm{mm}^3$ minimum volume. Physical units make the rule spacing-independent.

The anchor adds at most 20 TC boundary voxels above 0.40 and applies a stricter RC rule of $20~\mathrm{mm}^3$/0.70/0.85. Nested regions are repaired by
\[
\widehat Y_{\mathrm{TC}}\leftarrow\widehat Y_{\mathrm{TC}}\cup\widehat Y_{\mathrm{ET}},\qquad
\widehat Y_{\mathrm{WT}}\leftarrow\widehat Y_{\mathrm{WT}}\cup\widehat Y_{\mathrm{TC}}.
\]
RC receives priority through $\widehat Y_r\leftarrow\widehat Y_r\setminus\widehat Y_{\mathrm{RC}}$ for $r\in\{\mathrm{ET},\mathrm{TC},\mathrm{WT}\}$. Labels are assigned as RC, ET, residual TC, and residual WT (4, 3, 1, 2); this prediction anchors both refinements.

\subsection{Parent-Supported ET Refinement}

Because the anchor can omit weak ET components, the first refinement searches for supported additions. It forms 26-connected proposals from the union of $p^{\mathrm{XL}}_{\mathrm{ET}}$, $p^{\mathrm{M}}_{\mathrm{ET}}$, and $p^{\mathrm{L}}_{\mathrm{ET}}$ at 0.25. The learned operation is an LCv2 LightGBM component classifier trained by five-fold case-disjoint cross-fitting; parent support is deterministic, not a second classifier. LCv2 includes a soft LCv1 case probability and fixes 250 trees, learning rate 0.04, 15 leaves, minimum child size 20, row/feature subsampling 0.85, and L2 regularization 1.0. Its global cutoff 0.5497123599 maximizes recall over the full training-OOF score set under the anchor false-positive budget; it is not inner-fold nested. A two-of-three rule then requires trainer support at 0.25 for ET and its bounding-box TC/WT parents. Eligible proposals are added only to ET, and TC/WT expand to contain them. Existing anchor voxels are never removed, and global filters are not rerun.

\subsection{Utility-Gated ET Refinement}

The final refinement screens components of the three-trainer ET union that were omitted by, and remain disconnected from, the frozen parent-supported ET mask and align with the N01/RGv3 pool. Two LightGBM classifiers \cite{ke2017lightgbm} estimate lesion existence (any reference ET overlap) and geometry safety (overlap with component precision $\ge0.50$). Five-fold case-disjoint training uses 3,265 components/825 cases: existence has 1,072/2,193 positive/negative targets and geometry safety 1,032/2,233. The 274 components/103 cases in the final gate audit are a nested application-aligned subset (existence 186/88; geometry 185/89), not a separate training set. All 70 fixed inputs comprise morphology/location (19), trainer and fused probabilities (32), support (6), uncertainty (3), containment (4), rank/distance (4), and upstream scores (2). No automated feature selection or nested tuning was used. Both models fix 160 trees, learning rate 0.03, 15 leaves, depth 4, minimum child size 20, row/feature subsampling 0.85, L2 regularization 2.0, and seed 20260728. For candidate $C$, $q_C$ is exactly its upstream LCv2 probability (the stored \texttt{v2\_component\_probability}), and $(e_C,g_C)$ are its existence and geometry-safety probabilities. The gate decision is
\begin{equation}
\operatorname{gate}(C)=
\begin{cases}
\mathrm{accept}, & \min(q_C,e_C,g_C)\ge0.75,\\
\mathrm{reject}, & \min(e_C,g_C)<0.50,\\
\mathrm{abstain}, & \text{otherwise}.
\end{cases}
\label{eq:utilitygate}
\end{equation}
Only accepted components are added; final set operations enforce $\mathrm{ET}\subseteq\mathrm{TC}\subseteq\mathrm{WT}$ and leave RC unchanged. The 0.50 reject and 0.75 unanimous-accept thresholds are global conservative decision states over training-OOF probabilities, not fold-wise optimized cutoffs. Utility V4 is retained to reproduce the frozen challenge submission, but its practical role is optional because its aggregate incremental effect is small. Supplementary Tables S3--S5 give parameter provenance, the complete 70-input definitions, and the component-level gate audit.

\subsection{Development Selection and Ablation Protocol}

Screening varied schedules, architectures, initialization, lesion-aware objectives, and sampling, so it supports configuration selection rather than single-factor attribution. The final configuration followed comparison of 101 outputs on the same 179 cases without a preregistered scalar objective; Small F1, All F1, DSC, NSD, and HD95 were considered jointly. Main ablations hold cases and probabilities fixed and follow execution order. Component-model OOF analyses remain component-level evidence, not full-mask OOF validation.

\FloatBarrier
\section{Results}

\subsection{Development Evidence for Trainer Selection}
\label{sec:development-results}

Each of the 1,295 training cases is predicted only by its assigned held-out fold. ResEncM uses the ResEnc-M plan, Dice--BCE, the configured nnU-Net schedule, and checkpoint-best selection, yielding WT/TC/ET/RC voxel Dice of 0.7558/0.7427/0.7208/0.1427. Lesion-sensitive ResEncM uses the ResEnc-M plan, Dice--BCE with focal and asymmetric Tversky terms, configured 300 epochs, and checkpoint-best selection, yielding 0.7721/0.7564/0.7337/0.1728. ResEncXL uses the ResEnc-XL30GB plan, Dice--BCE, configured 1000 epochs, and checkpoint-best selection, yielding 0.7735/0.7639/0.7391/0.2583. These are full-volume nnU-Net voxel metrics rather than official lesion-wise scores, and the configurations are non-factorial comparisons. Supplementary Tables S1--S2 report the candidate inventory and complete trainer configuration in greater detail.

Under the screening definition, focal--Tversky fine-tuning increased micro-averaged component recall from 0.302 (442 TP, 1,022 FN) to 0.367 (538 TP, 926 FN), while precision changed from 0.585 to 0.584 as FP increased from 313 to 384. Micro-F1 consequently increased from 0.398 to 0.451. This sensitivity--precision trade-off motivates retaining the lesion-sensitive trainer as a probability source rather than using it as a standalone replacement. In the ResEncM-scale analysis, component filtering subsequently increased DSC/NSD from 0.665/0.672 to 0.700/0.706 and reduced HD95 from 72.3 to 58.5~mm, showing that lesion-sensitive probabilities and conservative component decisions address different errors.

\subsection{Overall System Comparison and Stagewise Ablation}

Table~\ref{tab:stagewise_results} connects the retained ResEncM trainer endpoints with ResEncXL and the deployed inference sequence on the same 179 cases. DSC, NSD, All F1, and HD95 are directly comparable across all rows. Small F1 is not cross-compared for the two ResEncM endpoints because their historical score files use the screening definition reported above; the ResEncXL and final-sequence rows use the official Task~1 definition.

\begin{table}[t]
\caption{Overall evidence ladder on 179 cases. DSC, NSD, All F1, and HD95 are directly comparable. Small F1 is omitted for the historical ResEncM endpoints because they use the screening definition reported above. Best values are bold.}
\label{tab:stagewise_results}
\centering\scriptsize
\setlength{\tabcolsep}{2.3pt}
\renewcommand{\arraystretch}{0.91}
\begin{tabular}{@{}L{0.42\linewidth}ccccc@{}}
\toprule
Configuration & Small F1$\uparrow$ & DSC$\uparrow$ & NSD$\uparrow$ & All F1$\uparrow$ & HD95$\downarrow$ \\
\midrule
\multicolumn{6}{@{}l}{\textit{Trainer endpoints}} \\
ResEncM 5-fold & \textemdash{} & 0.6645 & 0.6719 & 0.6712 & 72.35 \\
Lesion-sensitive ResEncM 5-fold & \textemdash{} & 0.6643 & 0.6747 & 0.6603 & 74.86 \\
ResEncXL 5-fold & \textbf{0.3666} & 0.6657 & 0.6780 & 0.7408 & 69.23 \\
\addlinespace[1.5pt]
\multicolumn{6}{@{}l}{\textit{Final inference sequence}} \\
Structured-probability anchor & 0.3193 & 0.7088 & 0.7130 & 0.7760 & 56.82 \\
+ Parent-supported ET & 0.3536 & 0.7120 & 0.7136 & 0.7789 & 56.66 \\
\textbf{+ Utility gate (COMPASS-METS)} & 0.3538 & \textbf{0.7121} & \textbf{0.7137} & \textbf{0.7790} & \textbf{56.64} \\
\bottomrule
\end{tabular}
\end{table}

The lesion-sensitive ResEncM endpoint slightly increased NSD relative to the standard ResEncM ensemble (0.6747 versus 0.6719), but reduced All F1 (0.6603 versus 0.6712) and increased HD95 (74.86 versus 72.35~mm). This error-profile shift supports its use as an auxiliary probability source rather than as a standalone replacement. ResEncXL increased All F1 to \XLBaseAllFOne{} and reduced HD95 to \XLBaseHD~mm. Within the final inference sequence, the structured anchor produced most of the geometric gain but reduced Small F1 to 0.3193. Parent-supported refinement restored Small F1 to 0.3536 while retaining the anchor's DSC, NSD, All F1, and HD95 gains. COMPASS-METS finally obtained \XLFinalSmallFOne, \XLFinalDSC, \XLFinalNSD, \XLFinalAllFOne, and \XLFinalHD~mm, respectively. An existing equal-logit XL/M/lesion-sensitive control followed by the common component gate obtained Small F1/DSC/NSD/All F1/HD95 of 0.312359/0.705126/0.708612/0.768333/57.844~mm. COMPASS-METS changed these by $+0.041435/+0.006929/+0.005070/+0.010641/-1.202$~mm. Because this control includes component gating, it does not replace pure averaging and majority-vote controls under a case-disjoint protocol.

\begin{table}[t]
\caption{Official 27~mm$^3$ lesion-wise post-selection held-out-fold comparison on 1,295 cases. Values are paired case-level macro averages across ET, RC, TC, and WT; best values are bold.}
\label{tab:heldout_oof}
\centering\scriptsize
\setlength{\tabcolsep}{2.1pt}
\renewcommand{\arraystretch}{0.91}
\begin{tabular}{@{}L{0.40\linewidth}ccccc@{}}
\toprule
Candidate & Small F1$\uparrow$ & DSC$\uparrow$ & NSD$\uparrow$ & All F1$\uparrow$ & HD95$\downarrow$ \\
\midrule
ResEncXL & 0.5948 & 0.6238 & 0.6641 & 0.7462 & 105.68 \\
Equal-probability average & 0.5631 & 0.6157 & 0.6567 & 0.7253 & 108.29 \\
Equal-logit average & 0.5719 & 0.6210 & 0.6620 & 0.7358 & 106.50 \\
Hard majority vote & 0.5531 & 0.6119 & 0.6528 & 0.7183 & 109.80 \\
Parent-supported COMPASS-METS & 0.6275 & 0.6546 & 0.6877 & 0.7691 & 99.86 \\
\textbf{Utility-V4 COMPASS-METS} & \textbf{0.6275} & \textbf{0.6546} & \textbf{0.6877} & \textbf{0.7691} & \textbf{99.85} \\
\bottomrule
\end{tabular}
\end{table}

\begin{figure}[t]
    \centering
    \includegraphics[width=0.70\linewidth]{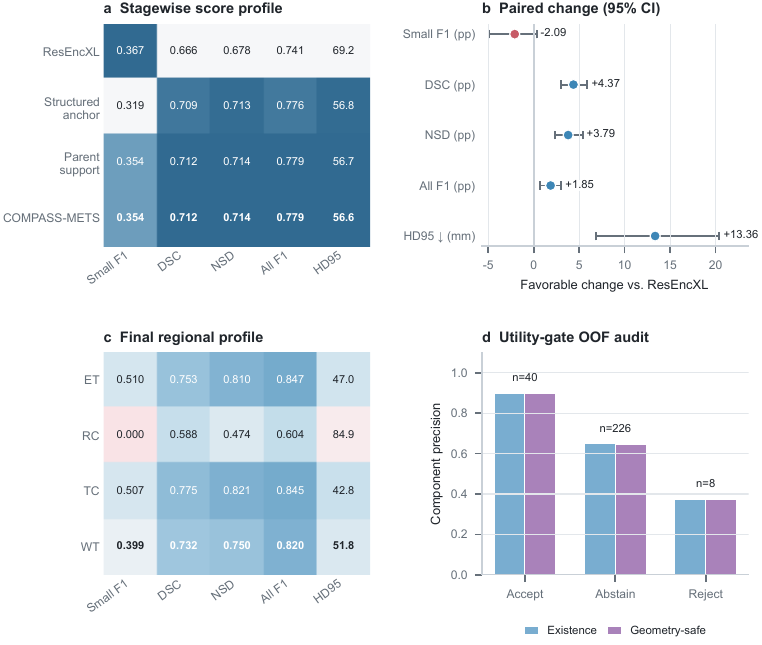}
    \caption{Quantitative analysis of COMPASS-METS: (a) stagewise scores, (b) paired COMPASS-METS-versus-ResEncXL changes with 95\% bootstrap intervals, (c) regional performance of the final system, and (d) component-level OOF precision of the three-state utility gate. Positive paired changes are favorable; HD95 is therefore shown as a reduction.}
    \label{fig:xlmresults}
\end{figure}

Across the 1,295 held-out-fold cases, Utility-V4 COMPASS-METS achieved Small F1/DSC/NSD/All F1 of 0.6275/0.6546/0.6877/0.7691 and HD95 of 99.85~mm (Table~\ref{tab:heldout_oof}). Relative to ResEncXL, the paired changes were $+0.0328$ Small F1 (95\% CI 0.0227 to 0.0434), $+0.0308$ DSC (0.0256 to 0.0362), $+0.0236$ NSD (0.0181 to 0.0293), $+0.0229$ All F1 (0.0168 to 0.0292), and a 5.82~mm HD95 reduction (3.56 to 8.21~mm). The favorable direction held in every fold for all five metrics (Supplementary Table~S8). None of the three pure ensemble controls exceeded ResEncXL on these case-macro metrics. Utility V4 and parent support were identical at four-decimal precision for the four unitless metrics, with a 0.004~mm HD95 difference, confirming that parent support accounts for essentially all of the OOF improvement. The paired COMPASS-METS-versus-ResEncXL analysis yielded a Small F1 change of \XLDeltaSmallFOne{} (95\% CI \XLDeltaSmallFOneLow{} to \XLDeltaSmallFOneHigh), DSC change of \XLDeltaDSC{} (\XLDeltaDSCLow{} to \XLDeltaDSCHigh), NSD change of \XLDeltaNSD{} (\XLDeltaNSDLow{} to \XLDeltaNSDHigh), and All F1 change of \XLDeltaAllFOne{} (\XLDeltaAllFOneLow{} to \XLDeltaAllFOneHigh). HD95 was reduced by \XLDeltaHD~mm (\XLDeltaHDLow{} to \XLDeltaHDHigh~mm). Figure~\ref{fig:xlmresults} summarizes the stagewise, paired, regional, and component-gate evidence.

\subsection{Regional and Component-Gate Analysis}

\begin{figure}[t]
    \centering
    \includegraphics[width=0.8\linewidth]{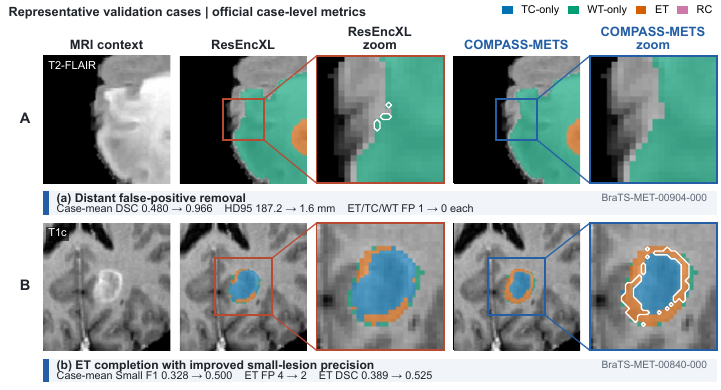}
    \caption{Evaluator-backed ResEncXL--COMPASS-METS development examples: (a) false-positive removal and (b) ET completion. Lines connect ROIs to enlargements; hidden reference masks are not displayed.}
    \label{fig:qualitative}
\end{figure}

ET achieved All F1/DSC/NSD of \XLETAllFOne/\XLETDSC/\XLETNSD, while TC and WT achieved \XLTCAllFOne/\XLTCDSC/\XLTCNSD and \XLWTAllFOne/\XLWTDSC/\XLWTNSD, respectively. RC remained the least certain region: All F1 was \XLRCAllFOne{} over 27 scorable cases; DSC, NSD, and HD95 were 0.588, 0.474, and 84.92~mm over 22 cases; and Small F1 was 0.000 over only six cases. These RC values therefore have high sampling uncertainty and are not interpreted as stable region-level estimates. Supplementary Tables S6--S7 give all regional denominators and evaluator-backed failure cases; Fig.~\ref{fig:qualitative} provides the qualitative comparison. Across 274 cross-fit candidates from 103 cases, accept ($n=40$), abstain ($n=226$), and reject ($n=8$) had existence precisions 0.900, 0.650, and 0.375 and geometry-safe precisions 0.900, 0.646, and 0.375. The accepted set included 36 candidates meeting both targets. On the challenge cohort, the gate changed one case by six ET voxels, adding two matched ET components and no false-positive component.

\paragraph{Computational requirements.}
A frozen Docker run processed 179 cases in 5~h 32~min 02~s on one RTX~3070 (8~GB) and eight CPU threads (111.3~s/case); the archive was 9.18~GB. One-GPU-per-fold logs sum to 1,091.220 epoch-duration GPU-hours, not calendar or billing time. Source/trimmed checkpoints occupy 11.418/5.708~GiB. The per-trainer hardware, batch-size, and logged-duration breakdown is given in the supplementary recorded-computation section.

\section{Discussion}

ResEncXL supplies primary predictions and lesion-sensitive ResEncM contributes alternative probabilities. The structured anchor delivers most geometric gains, parent support recovers most lost Small F1, and the 0.900-precision utility gate makes a narrow final adjustment (Fig.~\ref{fig:qualitative}). Limitations are development-cohort reuse, component-table OOF utility evidence, fixed proposals, non-factorial screening, unequal fine-tuning, low RC prevalence, and 15-model inference; results are therefore selection-aware estimates. This 15-model research prototype uses retrospective de-identified challenge data, lacks prospective/external/clinical validation, and should not guide care without governance and clinician oversight. The 1,295-case evaluation uses case-disjoint learned predictions but remains post-selection because operating points are global; priorities remain inner-fold operating-point selection, external/prospective validation, calibrated RC evaluation, and lower-cost deployment.

\section{Conclusion}

We introduced COMPASS-METS, which constructs aligned regional probabilities from three five-fold nnU-Net trainers and combines structured probability construction with parent-supported and utility-gated ET refinement. Across 179 challenge-development cases, it achieved Small F1/DSC/NSD/All F1 of \XLFinalSmallFOne/\XLFinalDSC/\XLFinalNSD/\XLFinalAllFOne{} and HD95 of \XLFinalHD~mm. The 1,295-case post-selection held-out-fold evaluation showed the same favorable direction versus ResEncXL across all five metrics, whereas the three pure ensemble controls did not outperform it. Stagewise analysis linked geometric gains to structured construction and most small-lesion recovery to parent support.

\begin{credits}
\subsubsection{\ackname} This work was supported by PolyU Start-up Fund P0060371.

\subsubsection{\discintname}
The authors have no competing interests to declare that are
relevant to the content of this article.
\end{credits}

\bibliographystyle{splncs04}
\bibliography{refs}

@misc{https://doi.org/10.5281/zenodo.19714728,
  doi = {10.5281/ZENODO.19714728},
  url = {https://zenodo.org/doi/10.5281/zenodo.19714728},
  author = {Bakas, Spyridon and Linguraru, Marius and Kazerooni, Anahita Fathi and Farahani, Keyvan and Astaraki, Mehdi and Maleki, Nazanin and Menze, Bjoern and Baid, Ujjwal and Yordanov, Nikolay and Kofler, Florian and You, Suhang and Chung, Verena and Mangul, Serghei and Velichko, Yury and Gandhi, Deep and Jiang, Zhifan and Conte, Gian Marco and Moawad, Ahmed and LaBella, Dominic and De Verdier, Maria Correia and Aboian, Mariam and Wiestler, Benedikt and Huse, Jason and Huang, Raymond},
  title = {{BraTS} 2026 Cluster of Challenges},
  publisher = {Zenodo},
  year = {2026},
  copyright = {Creative Commons Attribution 4.0 International}
}

@article{moawad2024brain,
  title={The Brain Tumor Segmentation--Metastases ({BraTS-METS}) Challenge 2023: Brain Metastasis Segmentation on Pre-treatment {MRI}},
  author={Moawad, Ahmed W and Janas, Anastasia and Baid, Ujjwal and Ramakrishnan, Divya and Saluja, Rachit and Ashraf, Nader and Maleki, Nazanin and Jekel, Leon and Yordanov, Nikolay and Fehringer, Pascal and others},
  journal={arXiv preprint arXiv:2306.00838},
  year={2024}
}

@article{maleki2025analysis,
  title={Analysis of the {MICCAI} Brain Tumor Segmentation--Metastases ({BraTS-METS}) 2025 Lighthouse Challenge: Brain Metastasis Segmentation on Pre- and Post-treatment {MRI}},
  author={Maleki, Nazanin and Amiruddin, Raisa and Moawad, Ahmed W and Yordanov, Nikolay and Gkampenis, Athanasios and Fehringer, Pascal and Umeh, Fabian and Chukwurah, Crystal and Memon, Fatima and Petrovic, Bojan and others},
  journal={arXiv preprint arXiv:2504.12527v3},
  year={2025}
}

@article{cho2021brain,
  title={Brain metastasis detection using machine learning: a systematic review and meta-analysis},
  author={Cho, Se Jin and Sunwoo, Leonard and Baik, Sung Hyun and Bae, Yun Jung and Choi, Byung Se and Kim, Jae Hyoung},
  journal={Neuro-oncology},
  volume={23},
  number={2},
  pages={214--225},
  year={2021},
  publisher={Oxford University Press US}
}

@article{ambrosini2010computer,
  title={Computer-aided detection of metastatic brain tumors using automated three-dimensional template matching},
  author={Ambrosini, Robert D and Wang, Peng and O'Dell, Walter G},
  journal={Journal of Magnetic Resonance Imaging},
  volume={31},
  number={1},
  pages={85--93},
  year={2010},
  publisher={Wiley Online Library}
}

@article{isensee2021nnu,
  title={{nnU-Net}: A Self-configuring Method for Deep Learning-based Biomedical Image Segmentation},
  author={Isensee, Fabian and Jaeger, Paul F and Kohl, Simon AA and Petersen, Jens and Maier-Hein, Klaus H},
  journal={Nature methods},
  volume={18},
  number={2},
  pages={203--211},
  year={2021},
  publisher={Nature Publishing Group US New York}
}

@inproceedings{isensee2023extending,
  title={Extending {nnU-Net} Is All You Need},
  author={Isensee, Fabian and Ulrich, Constantin and Wald, Tassilo and Maier-Hein, Klaus H},
  booktitle={BVM workshop},
  pages={12--17},
  year={2023},
  organization={Springer}
}

@inproceedings{ke2017lightgbm,
  title={{LightGBM}: A Highly Efficient Gradient Boosting Decision Tree},
  author={Ke, Guolin and Meng, Qi and Finley, Thomas and Wang, Taifeng and Chen, Wei and Ma, Weidong and Ye, Qiwei and Liu, Tie-Yan},
  booktitle={Advances in Neural Information Processing Systems},
  volume={30},
  year={2017}
}

@article{karargyris2023federated,
  title={Federated benchmarking of medical artificial intelligence with MedPerf},
  author={Karargyris, Alexandros and Umeton, Renato and Sheller, Micah J and Aristizabal, Alejandro and George, Johnu and Wuest, Anna and Pati, Sarthak and Kassem, Hasan and Zenk, Maximilian and Baid, Ujjwal and others},
  journal={Nature machine intelligence},
  volume={5},
  number={7},
  pages={799--810},
  year={2023},
  publisher={Nature Publishing Group UK London}
}

@article{luo2025msm,
  title={MSM-Seg: A Modality-and-Slice Memory Framework with Category-Agnostic Prompting for Multi-Modal Brain Tumor Segmentation},
  author={Luo, Yuxiang and Xu, Qing and Huang, Hai and Ouyang, Yuqi and Chen, Zhen and Duan, Wenting},
  journal={arXiv preprint arXiv:2510.10679},
  year={2025}
}

@article{xu2023dcsau,
  title={DCSAU-Net: A deeper and more compact split-attention U-Net for medical image segmentation},
  author={Xu, Qing and Ma, Zhicheng and Duan, Wenting and others},
  journal={Computers in biology and medicine},
  volume={154},
  pages={106626},
  year={2023},
  publisher={Elsevier}
}

@article{liu2025harnessing,
  title={Harnessing lightweight transformer with contextual synergic enhancement for efficient 3d medical image segmentation},
  author={Liu, Xinyu and Chen, Zhen and Li, Wuyang and Li, Chenxin and Yuan, Yixuan},
  journal={IEEE Transactions on Pattern Analysis and Machine Intelligence},
  year={2025},
  publisher={IEEE}
}

\end{document}